\documentclass[sigconf,natbib=true,anonymous=false,review=false]{acmart}

\copyrightyear{2025}
\acmYear{2025}
\acmConference[ICTIR '25]{Proceedings of the 2025 International ACM SIGIR Conference on Innovative Concepts and Theories in Information Retrieval (ICTIR)}{July 18, 2025}{Padua, Italy}
\acmBooktitle{Proceedings of the 2025 International ACM SIGIR Conference on Innovative Concepts and Theories in Information Retrieval (ICTIR) (ICTIR '25), July 18, 2025, Padua, Italy}\acmDOI{10.1145/3731120.3744595}
\acmISBN{979-8-4007-1861-8/2025/07}

\usepackage{enumitem}
\newlist{inlinelist}{enumerate*}{1}
\setlist*[inlinelist,1]{%
  label=(\roman*),
}
\usepackage{subfigure}
\usepackage{xcolor}
\usepackage{xspace}
\usepackage{multirow}
\usepackage{adjustbox}
\usepackage{array}
\newcolumntype{H}{>{\setbox0=\hbox\bgroup}c<{\egroup}@{}}
\usepackage{booktabs}
\usepackage{microtype}
\usepackage{graphicx}

\title{Comparing Retrieval-Augmentation and Parameter-Efficient Fine-Tuning for Privacy-Preserving Personalization of Large Language Models}

\author{Alireza Salemi}
\affiliation{\institution{University of Massachusetts Amherst}
\city{Amherst}
\state{MA}
\country{United States}}
\email{asalemi@cs.umass.edu}

\author{Hamed Zamani}
\affiliation{\institution{University of Massachusetts Amherst}
\city{Amherst}
\state{MA}
\country{United States}}
\email{zamani@cs.umass.edu}

\begin{document}


\begin{abstract}
Despite its substantial impact on various search, recommendation, and question answering tasks, privacy-preserving methods for personalizing large language models (LLMs) have received relatively limited exploration. There is one primary approach in this area through retrieval-augmented generation (RAG), which generates personalized outputs by enriching the input prompt with information retrieved from the user’s personal data. This paper studies an orthogonal approach to RAG that involves learning user-dependent LLM parameters through parameter-efficient fine-tuning (PEFT). This paper presents the first systematic study for exploration of PEFT for LLM personalization and provides an extensive comparisons between RAG- and PEFT-based solutions, across a broad set of seven diverse datasets from the LaMP benchmark. Our results demonstrate that, on average, both RAG- and PEFT-based personalization methods yield 14.92\% and 1.07\% improvements over non-personalized LLMs, respectively. When combining RAG with PEFT, we observe a further improvement of 15.98\%, highlighting the effectiveness of their integration in enhancing personalized text generation. Additionally, we identify a positive correlation between the amount of user data available and the effectiveness of PEFT. This finding suggests that RAG is particularly beneficial for cold-start users---users with limited personal data---while PEFT performs better when more user-specific data is available. 
\end{abstract}



\maketitle

\section{Introduction}


Personalization has been a longstanding area of interest in the information retrieval community and has been extensively studied for tasks such as search \cite{10.1162/dint_a_00104}, recommendation \cite{10.1145/2766462.2767707, 10.1145/1454008.1454048, 10386507}, question answering \cite{10.1145/3589335.3651445}, and conversational search \cite{cho-etal-2021-personalized, 10.1007/978-3-030-35330-8_4}. With the rise of LLMs, personalizing them has recently emerged as a critical topic \cite{lamp, rspg, longlamp} due to its applications in various real-world systems, such as personalized recommender systems \cite{llm-rs-tutorial, chen2023surveylargelanguagemodels}, virtual assistants \cite{li2024helloagainllmpoweredpersonalized, info:doi/10.2196/15360}, and content generation \cite{alhafni-etal-2024-personalized}. These systems benefit from tailoring responses and actions based on individual user preferences, leading to enhanced user experiences, greater user satisfaction, and more effective interactions \cite{WALKER2004811}. Personalization allows LLMs to better understand and predict user needs, offering more relevant and contextually appropriate content or responses \cite{lamp}. This ability to adjust to the unique characteristics of individual users, such as their interests, behaviors, and prior interactions, is what sets personalized systems apart from general-purpose models.

Retrieval-augmented generation (RAG) \cite{rag, reml} has emerged as an effective approach for enhancing various aspects of machine learning tasks, particularly in text generation. RAG \cite{rag, fid} integrates information retrieval with natural language generation, thereby improving the relevance and factual accuracy of the generated outputs. Unlike traditional large language models (LLMs), which rely solely on the static knowledge obtained during pre-training, RAG systems utilize a retriever to access external information at inference time, enabling contextually grounded and factually consistent generation \cite{asai2023selfrag, siriwardhana-etal-2023-improving}. This capability allows RAG to produce more informed and relevant outputs by leveraging up-to-date or personal data. Currently, the most reliable approach to personalizing LLMs is through RAG personalization, where personalized information is retrieved from the user's profile to construct a personalized input prompt for the LLM \cite{lamp, rspg}. This personalized prompt provides the LLM with relevant user context, enabling it to generate more tailored and contextually appropriate responses. This approach leverages the personalization capabilities of retrieval models to enhance the LLM's output relevance. 

Another potential approach to personalizing LLMs involves fine-tuning the LLM on user-specific data, such as documents authored by the user or other relevant information about the user. However, training a separate LLM for each user is computationally expensive and requires substantial storage resources, preventing scalability of this approach. To solve this, parameter-efficient fine-tuning (PEFT) techniques, such as low-rank adaptation (LoRA) \cite{hu2022lora}, can be used to adjust a subset of the model's parameters based on each user's data. LoRA has demonstrated significant reductions in memory consumption and storage requirements while increasing performance for tasks such as language modeling, reasoning, text generation, and question answering \citep{zhao2024loraland310finetuned, zhang2023lorafamemoryefficientlowrankadaptation, pan2024lisa}. Thus, using LoRA to train the LLM on individual user data personalizes the LLM while avoiding the need to store a full set of parameters for each user, which would be computationally expensive in large-scale applications. Given the success story of PEFT in the mentioned applications, this paper presents a study of a established PEFT method for LLM personalization, by selectively tuning the parameters with user-specific data. This way PEFT can potentially align the model's knowledge and behavior with the user's preferences and requirements. Furthermore, RAG- and PEFT-based approaches can be integrated to enhance personalization, leveraging the strengths of both RAG and fine-tuning techniques for more effective user adaptation. 

A key concern in personalization of LLMs is that using private user data to train LLMs may compromise user privacy. This risk is heightened when a shared model is used across multiple users and learns from all of their data. To address this, we focus on methods that use and train LLMs exclusively on user-specific data, ensuring that no data is shared between models for different users. Under this framework, both methods for personalizing LLMs ---RAG and PEFT---maintain the user's privacy, as they neither update model nor generate input prompts using data from other users. Even though the LLM leverages private personal information to generate more tailored responses for the user, this data is only accessible to the user themselves. Since the information is not shared with other users or exposed to the model in a way that would affect other users' data, this approach does not constitute a privacy violation.

To systematically study these two schools of thoughts for personalizing LLMs, this paper studies a PEFT method for LLM personalization and compares it against more established RAG-based solutions in a privacy-preserving setting. Specifically, we focus on utilizing PEFT methods for personalizing LLMs, as this is an underexplored avenue in this area. To achieve this, we conduct an extensive set of experiments using seven diverse datasets from the Language Model Personalization (LaMP) benchmark \cite{lamp}. LaMP comprises three classification tasks and four text generation tasks, each designed to evaluate different aspects of LLM personalization. In this benchmark, each input is treated as a separate user, with its own distinct input, expected output, and user profile. This structure makes LaMP an ideal testbed for evaluating the effectiveness of the personalization methods explored in this study. We address the following research questions to compare these two methods:
\begin{itemize}[leftmargin=*, itemsep=1em]
    \item \textbf{RQ1: How do PEFT- and RAG-based approaches perform for LLM personalization?} Our experiments reveal that personalizing LLMs using retrieval-augmented generation results in an average improvement of 14.92\% over the non-personalized baseline, while parameter-efficient fine-tuning based personalization leads to only a modest 1.07\% improvement.

    \item \textbf{RQ2: How does the combination of PEFT and RAG impact the personalization performance?} Our results show that combining both RAG and PEFT yields the best overall performance on personalized tasks, achieving a 15.98\% improvement over the non-personalized baseline. This suggests that integrating both approaches is the most effective strategy for personalizing LLMs.

    \item \textbf{RQ3: How does profile size affect performance?} To deepen our understanding, we provide an analysis to explain why PEFT does not perform as well for personalizing LLMs. Our findings indicate a positive correlation between the size of the user profile and performance improvement, suggesting that the limited data per user is a key factor contributing to PEFT's underperformance.

    \item \textbf{RQ4: How does data presence in training corpus affect performance?} We observe that improvements from PEFT are most significant when the dataset used for training is not publicly accessible and is unlikely to be used in the pre-training data. Notably, PEFT achieves the highest gains on the Personalized Email Subject Generation task, which is based on a private dataset.
\end{itemize}

Finally, we discuss the limitations of each approach to provide a clearer comparison. For instance, while PEFT is more resource-intensive than RAG due to the need for fine-tuning and storing user-specific parameters, it is generally faster at inference time since it does not require retrieving external information. In contrast, RAG is more efficient in terms of storage and adaptation to new information but incurs additional latency during retrieval. By examining these trade-offs, we provide a comprehensive evaluation of each method's advantages and limitations. To promote further exploration of this important area, we will open-source our implementation and integrate it with the LaMP codebase, ensuring that researchers can build upon and validate our findings.\footnote{The code and data used in our experiments in this paper are available at: \url{https://github.com/LaMP-Benchmark/LaMP}}

\section{Related Work}

\subsection{\textbf{Retrieval-Augmented Generation}}

RAG \cite{rag, fid} is a framework that enhances natural language generation by integrating information retrieval with the generation process, improving the relevance and factual accuracy of the generated content. Unlike traditional large language models (LLMs), which rely solely on the knowledge acquired during pre-training, RAG systems retrieve external information via a retriever to incorporate contextually grounded and factually consistent knowledge during inference \cite{asai2023selfrag, siriwardhana-etal-2023-improving}. This integration allows RAG to generate outputs that are more informed and relevant by leveraging real-time external data. The versatility of RAG spans several domains, including knowledge-grounded text generation \cite{kilt, rag, fid}, multi-modal reasoning \cite{dedr, murag, pretrain-kivqa}, personalized generation \cite{lamp, rspg}, and mitigating hallucinations in model outputs \cite{shuster-etal-2021-retrieval-augmentation}. This adaptability makes RAG a powerful tool for applications that require up-to-date or user-specific information. In this paper, we employ RAG as one of the methods for personalizing LLMs, following the approach presented by \citet{lamp}, which is one of the leading methods in the field of personalized LLMs.

\subsection{\textbf{Parameter Efficient Fine-Tuning}}

PEFT enables the adaptation of large language models (LLMs) to specific tasks without the need for full model retraining, significantly reducing computational costs while maintaining strong performance \cite{pmlr-v97-houlsby19a, liao-etal-2023-parameter, han2024parameterefficient}. Among various PEFT methods, Low-Rank Adaptation (LoRA) is particularly effective for efficiently fine-tuning LLMs. It introduces low-rank decomposition into weight matrices and injects trainable low-rank matrices into otherwise frozen model weights \cite{hu2022lora, li2024loftq, lialin2024relora}. This technique minimizes the number of trainable parameters while preserving the model’s expressiveness, making it a scalable and computationally efficient alternative to traditional full fine-tuning. Due to its ability to significantly reduce the cost of model training, PEFT has gained attention for personalizing LLMs, allowing models to adapt to diverse user needs with relatively low resource consumption. However, prior work has typically maintained a shared pool of adapters across users, rather than training a separate adapter for each individual user \cite{tan2024personalized}. This shared-adapter approach, while efficient, introduces potential risks of data leakage and privacy concerns, as information from different users may inadvertently influence the model’s behavior. This paper investigates the use of LoRA in a privacy-preserving manner by training adapters exclusively on each user's data. Additionally, we explore the effectiveness of retrieval-augmented generation for LLM personalization. By comparing the two methods, this study examines their individual contributions to personalizing LLMs and explores potential synergies between PEFT and RAG, offering insights into optimizing LLM personalization through both parameter-efficient fine-tuning and retrieval-based augmentation while ensuring privacy protection.

\subsection{\textbf{Personalizing LLMs}}

Personalizing LLMs is a critical research area with applications in search, recommendation, and text generation \cite{10.1145/2702123.2702503, 10.1145/1462198.1462203, naumov2019deep, lamp}. \citet{lamp} introduced a retrieval-augmented generation based approach for personalizing LLMs and proposed the LaMP benchmark as a framework for evaluating personalized text generation. Another line of research has focused on developing personalized writing assistants \cite{li2023teach, mysore2023pearl, lu2024corporate} and autonomous agents \cite{zhang-etal-2024-llm-based}. Various approaches for LLM personalization have been proposed, including training retrieval models based on user feedback for text generation \cite{rspg}, optimizing LLMs with personalized feedback \cite{jang2023personalized}, and generating personalized prompts automatically \cite{Li_2024}. In parallel, recent studies have explored parameter-efficient fine-tuning \cite{tan2024personalized}, where a shared pool of adapters is trained on data from multiple users to personalize LLMs. However, this approach raises significant privacy concerns, as it may inadvertently lead to data leakage between users, compromising sensitive information. To address these concerns, this paper compares two prominent methods for personalizing LLMs---RAG and PEFT---in a privacy-preserving manner. Specifically, we explore training the LLM separately on each user's data to ensure privacy, while maintaining the model's ability to personalize responses effectively. This study seeks to enhance LLM personalization techniques while safeguarding user privacy by limiting model updates to user-specific data.

\begin{table*}[!t]
    \centering
    \caption{Prompts template used to augment the input of the LM with the user profile. We follow previous work by \citet{lamp} to implement these functions. \textcolor{blue}{\texttt{concat}} is a function that \texttt{concat}enates the strings in its first argument by placing the string in the second argument between them. \textcolor{blue}{\texttt{add\_to\_paper\_title}} is a function designed to add the string in its first argument to the paper's title in the Personalized Citation Identification task. \textcolor{blue}{\texttt{PPEP}} is a function that create the prompt for each entry in the retrieved profile entries. \textcolor{red}{\texttt{[INPUT]}} is the task's input.}
    \begin{adjustbox}{max width=\textwidth}
    \begin{tabular}{p{4cm}|p{7cm}|p{9cm}}
        \toprule
        \textbf{Task} & \textbf{Per Profile Entry Prompt (\texttt{PPEP})} & \textbf{Aggregated Input Prompt(AIP)} \\
        \midrule
            1: Citation Identification & ``$P_i$\texttt{\texttt{[title]}}'' & \textcolor{blue}{\texttt{add\_to\_paper\_title}}(\textcolor{blue}{\texttt{concat}}([\textcolor{blue}{\texttt{PPEP}}($P_1$), ..., \textcolor{blue}{\texttt{PPEP}}($P_n$)], \textcolor{gray}{", and "}), \textcolor{red}{\texttt{[INPUT]}}) \\ \midrule
            2: Movie Tagging & the tag for the movie: ``$P_i$\texttt{[description]}'' is ``$P_i$\texttt{[tag]}'' & \textcolor{blue}{\texttt{concat}}([\textcolor{blue}{\texttt{PPEP}}($P_1$), ..., \textcolor{blue}{\texttt{PPEP}}($P_n$)], \textcolor{gray}{``, and ''}). \textcolor{red}{\texttt{[INPUT]}} \\ \midrule
            3: Product Rating & $P_i$[score] is the score for ``$P_i$\texttt{[text]}'' & \textcolor{blue}{\texttt{concat}}([\textcolor{blue}{\texttt{PPEP}}($P_1$), ..., \textcolor{blue}{\texttt{PPEP}}($P_n$)], \textcolor{gray}{``, and ''}). \textcolor{red}{\texttt{[INPUT]}} \\ \midrule
            4: News Headline Generation & ``$P_i$\texttt{[title]}'' is the title for ``$P_i$\texttt{[text]}'' & \textcolor{blue}{\texttt{concat}}([\textcolor{blue}{\texttt{PPEP}}($P_1$), ..., \textcolor{blue}{\texttt{PPEP}}($P_n$)], \textcolor{gray}{``, and ''}). \textcolor{red}{\texttt{[INPUT]}} \\ \midrule
            5: Scholarly Title Generation & ``$P_i$\texttt{[title]}'' is the title for ``$P_i$[abstract]'' & \textcolor{blue}{\texttt{concat}}([\textcolor{blue}{\texttt{PPEP}}($P_1$), ..., \textcolor{blue}{\texttt{PPEP}}($P_n$)], \textcolor{gray}{``, and ""})\textcolor{gray}{. Following the given patterns} \textcolor{red}{\texttt{[INPUT]}} \\ \midrule
            6: Email Subject Generation & ``$P_i$\texttt{[title]}'' is the title for ``$P_i$\texttt{[text]}'' & \textcolor{blue}{\texttt{concat}}([\textcolor{blue}{\texttt{PPEP}}($P_1$), ..., \textcolor{blue}{\texttt{PPEP}}($P_n$)], \textcolor{gray}{``, and ''}). \textcolor{red}{\texttt{[INPUT]}} \\ \midrule
            7: Tweet Paraphrasing & ``$P_i$\texttt{[text]}'' & \textcolor{blue}{\texttt{concat}}([\textcolor{blue}{\texttt{PPEP}}($P_1$), ..., \textcolor{blue}{\texttt{PPEP}}($P_n$)], \textcolor{gray}{``, and ''}) \textcolor{gray}{are written by a person. Following the given patterns} \textcolor{red}{\texttt{[INPUT]}}\\
            \bottomrule
    \end{tabular}
    \end{adjustbox}
    
    \label{tab:task-prompts}
\end{table*}

\section{{Problem Formulation}}

A language model \( M \) takes a prompt \( x \) as input and generates an output \( \hat{y} \). However, the generated output is typically general, relying on the knowledge embedded in the model. To generate a personalized response, it is possible to incorporate a set of personalized information \( P_i \), also known as the user profile, which provides tailored details about the user to the LLM. This allows the model to produce responses that are more specific to the user's context and preferences. This paper focuses on personalized text generation, aiming to produce outputs that are tailored to the preferences of a user. We assume access to a dataset $T = \{(x_i, y_i, P_i)\}_{i=1}^{|T|}$, where $x_i$ is the input prompt from user $u_i$, $y_i$ is the expected output for user $u_i$, and $P_i$ is the user profile. Here, a user profile $P_i$ consists of a set of structured or unstructured text documents for the user $u_i$, denoted as $P_i = \{d_{(i, j)}\}_{j = 1}^{|P_i|}$. Note that $d_{(i, j)}$ can be either structured or unstructured documents, including input/output examples for the user or textual documents written by the user.

In this paper, we aim to leverage the user-specific information---provided by the user or collected through interactions with the user---available in the profile \( P_i \) to construct a personalized LLM \( M_{i} = \textsc{personalize}(M, P_i) \), by applying a transformation function \( \textsc{personalize} \) to the original LLM \( M \). This function can either modify the parameters of \( M \) to create \( M_i \) or alter the input to the LLM based on the profile \( P_i \). We focus on comparing different methods for designing the transformation \( \textsc{personalize} \) while maintaining privacy. Specifically, privacy is preserved by ensuring that no information from other users is used to personalize the LLM for a given user. A privacy violation occurs if the LLM reveals confidential information from one user to another. In this paper, we apply methods that work solely on each user's data, ensuring that the information is only revealed to the corresponding user, thus preventing any information leakage or privacy violations.


\section{Methods for Personalizing LLMs}

There are two main approaches to personalizing LLMs. The first approach involves personalizing the input prompt provided to the model, which encourages the model to generate responses that are more tailored to the specific user. The second approach focuses on changing the parameters of the LLM through training the LLM using user-specific data to help the model learn the user's preferences, writing style, and background knowledge, enabling it to generate responses that are better aligned with the individual user's needs and expectations. A major concern in personalizing LLMs is the potential for privacy violations. If the language model utilizes personal information from a user, there is a risk that sensitive data could be inadvertently revealed, compromising the user's privacy. This concern becomes particularly significant when information from one user is used to train a shared model, which may be accessed by multiple users. In such cases, there is a possibility that private information from one user could be leaked to others, violating privacy and undermining trust in the system. 

One straightforward approach to mitigate privacy risks in personalizing LLMs is to ensure that a personalized LLM for a specific user is exclusively used by that user. In this setup, the model can only access that user's information during both the training and inference phases. By restricting access to the user's personal data, this approach effectively eliminates the possibility of revealing sensitive information to other users, thus maintaining privacy and minimizing the risk of data leakage. With this in mind, this paper focuses on studying and investigating how each of the two introduced approaches contributes to personalizing LLMs in a fully privacy-preserving context. Specifically, we examine scenarios where the data of one user cannot be used to optimize the LLM for other users, ensuring that each user's personal information remains isolated and protected throughout the personalization process. The following sections provide a detailed explanation of these approaches.

\subsection{RAG for Personalizing LLMs}
\label{sec:rag}

In this approach, we employ a retrieval model to retrieve a set of personalized information from the user profile and use this information to construct a personalized prompt. The personalized prompt, which incorporates the user’s data, is then fed to the LLM to generate a tailored response. This method ensures that the output is specific to the individual user's context and preferences. 

Formally, given an input prompt $x$ from the user, we use the query generation function $\phi_q$ to create a query. This query is then passed through the retriever $R$, which retrieves $k$ documents from the user's profile $P_u$. Finally, the prompt generation function $\phi_p$ combines the retrieved documents and the input prompt to generate a personalized prompt, which is used as the input to the LLM $M$ to generate a more tailored response, formally, defined as: 
\begin{equation}
    \label{eq:general-rag}
    \hat{y} = M(\phi_{p}(x, R(\phi_{q}(x), k))
\end{equation}
where to implement the query generation function $\phi_q$, following \citet{lamp}, we extract and use the non-template portions of the user's input prompt as the query. For further details on the template used for generating inputs in the LaMP benchmark, we refer the reader to \citet{lamp}. Furthermore, we use the same function ($\phi_p$) as \citet{lamp} to generate personalized prompts for the LLM, as detailed in Table \ref{tab:task-prompts}.

Note that this approach does not modify the LLM itself. Instead, it adjusts its input, using a tailored prompt to the user based on the retrieved documents from the user profile. This allows us to personalize the LLM's response without altering its underlying structure and parameters, which works on any black-box LLM. In our experiments, we used a wide range of retrieval models: BM25 \cite{bm25} as a lexical-matching retrieval model, Contriever \cite{contriever} as a semantic matching retrieval model, Recency \cite{lamp} as a time-aware retrieval model, and RSPG \cite{rspg} as an ensemble model that chooses an appropriate retrieval model per input. This model ensures that each input is directed to the most suitable retrieval model for retrieving relevant documents that align with the given input. By selecting the optimal retrieval model, the system improves the relevance and accuracy of the retrieved information, thereby enhancing the overall response generation.

\begin{table*}
    \centering
    \caption{Implementation of the input-output generation function \( \textsc{convert} \) for PEFT personalization. The profiles in LaMP-2, LaMP-3, LaMP-4, LaMP-5, and LaMP-7 consist of input-output pairs, which are directly used as training pairs. However, for the LaMP-1 and LaMP-7, such pairs do not exist. For LaMP-1, we provide the model with a title and ask it to generate the abstract. For LaMP-7, we randomly divide a tweet into two parts and ask the model to generate the second part based on the prefix.}
    \adjustbox{max width=\textwidth}{\begin{tabular}{l|c|p{10cm}|c}
        \toprule
        {\textbf{Dataset}} & {\textbf{Profile Format}} & {\textbf{Generated Input ($x_i$)}} & {\textbf{Generated Output ($y_i$)}} \\
        \midrule

        \multirow{2}{*}{\shortstack[l]{{LaMP-1: Personalized}\\{Citation Identification}}} & title: [title] & \multirow{2}{*}{Write an abstract for this title: [title]} & \multirow{2}{*}{[abstract]} \\
        & abstract: [abstract] & & \\\midrule

        \multirow{6}{*}{\shortstack[l]{{LaMP-2: Personalized}\\{Movie Tagging}}} & \multirow{6}{*}{\shortstack[c]{{description: [description]} \\ {tag: [tag]}}} & \multirow{6}{*}{\shortstack[l]{{Which tag does this movie relate to among the following tags?} \\ {Just answer with the tag name without further explanation.} \\ {tags: [sci-fi, based on a book, comedy, action, twist ending,} \\ {dystopia, dark comedy, classic, psychology, fantasy, romance, } \\ {thought-provoking, social commentary, violence, true story]} \\ {description: [description]}}} & \multirow{6}{*}{[tag]} \\
        & & & \\ & & \\ & &\\ & & \\ & &\\\midrule

        \multirow{3}{*}{\shortstack[l]{{LaMP-3: Personalized}\\{Product Rating}}} & \multirow{3}{*}{\shortstack[l]{{review: [review]}\\{score: [score]}}} & \multirow{3}{*}{\shortstack[l]{{What is the score of the following review on a scale of 1 to 5?} \\ {just answer with 1, 2, 3, 4, or 5 without further explanation. }\\{review: [review]}}} & \multirow{3}{*}{[score]} \\ 
        & & \\ & & \\\midrule

        \multirow{2}{*}{\shortstack[l]{{LaMP-4: Personalized}\\{News Headline Generation}}} & article: [article] & \multirow{2}{*}{Generate a headline for the following article: [article]} & \multirow{2}{*}{[title]} \\
        & title: [title] & \\\midrule

        \multirow{2}{*}{\shortstack[l]{{LaMP-5: Personalized}\\{Scholarly Title Generation}}} & abstract: [abstract] & \multirow{2}{*}{Generate a title for the following abstract of a paper: [abstract]} & \multirow{2}{*}{[title]} \\
        & title: [title] & \\\midrule

        \multirow{2}{*}{\shortstack[l]{{LaMP-6: Personalized}\\{Email Subject Generation}}} & email: [email] & \multirow{2}{*}{Generate a subject for the following email: [email]} & \multirow{2}{*}{[title]} \\
        & title: [title] & \\\midrule

        \multirow{2}{*}{\shortstack[l]{{LaMP-7: Personalized}\\{Tweet Paraphrasing}}} & \multirow{2}{*}{tweet: [tweet]} & \multirow{2}{*}{Complete the following tweet: [first part of the tweet]} & \multirow{2}{*}{[second part of the tweet]} \\
        & & \\ \bottomrule
        
    \end{tabular}}
    
    \label{tab:imp-convert}
\end{table*}

\subsection{PEFT for Personalizing LLMs}
\label{sec:peft}

Maintaining a separate LLM for each user is impractical for systems with large user bases. For instance, storing a model like FlanT5-XXL \cite{flant5} requires 45 GB per user. With 1 million users, this amounts to 45,000 TB of storage, which is nearly infeasible for most systems to support. In addition to storage challenges, serving the full set of LLM parameters for each user requires substantial computational resources that cannot be efficiently shared across users. In contrast, a LoRA adapter with \( r = 8 \) for the same model only requires 55 MB of storage. For 1 million users, this would total 55 TB, which is far more manageable for real-world applications. Moreover, by using the same LLM backbone for all users and loading individual LoRA adapters per user, the system can operate more efficiently from a computational standpoint. Therefore, using Parameter-Efficient Fine-Tuning provides a more cost-effective solution compared to training an entire LLM for each user.

This approach uses a user profile \( P_u \) to learn user-specific parameters, resulting in a personalized LLM \( M_u \). There are various ways to achieve this personalization, and in our method, we apply LoRA (Low-Rank Adaptation) to the LLM \( M \) and train the model using the documents from \( P_u \). LoRA fine-tunes LLMs by injecting trainable low-rank matrices into the model’s weight matrices. Instead of updating all the model weights during training, LoRA decomposes the weight update into two smaller, low-rank matrices \( A \in \mathbb{R}^{d \times r} \) and \( B \in \mathbb{R}^{r \times k} \), where \( d \) is the dimension of the input, \( k \) is the dimension of the output, and \( r \) is the rank parameter that controls the capacity of the low-rank approximation. The original weight matrix \( W_0 \in \mathbb{R}^{d \times k} \) is kept frozen, and only the matrices \( A \) and \( B \) are trained. The low-rank matrices \( A \) and \( B \) are optimized to approximate the update to the original weights, such that the updated weights are \( W = W_0 + AB \). This allows LoRA to efficiently adapt the model by learning a compact set of parameters, without requiring the full weight matrix to be updated, thereby reducing computational and memory costs. The parameter \( r \), also known as the rank, is a crucial aspect of LoRA’s capacity. A larger \( r \) allows for more complex adaptations, while a smaller \( r \) ensures a more compact and efficient adaptation, but with potentially less expressiveness. In practice, \( r \) is chosen based on a trade-off between the computational efficiency (smaller \( r \)) and the model’s capacity to learn detailed user-specific adjustments (larger \( r \)). This low-rank adaptation approach not only ensures that the model can be personalized effectively for each user but also makes the process more computationally efficient, as it avoids the need to retrain or store a full set of model parameters for each user. Instead, only the smaller adaptation matrices \( A \) and \( B \) are learned and stored, making it feasible to personalize the model for a large number of users without excessive resource consumption.

To train the LLM using LoRA on a user profile \( P_u \), each document \( d \in P_u \) is first transformed into a pair of input-output sequences, where the LLM receives the input and generates the corresponding output. We define this transformation process using the function \( (x_i, y_i) = \textsc{convert}(d_i) \), where \( x_i \) represents the input sequence derived from \( d_i \) and \( y_i \) is the target output. For each document \( d_i \) in \( P_u \), we apply the conversion function to obtain the input-output pair \( (x_i, y_i) \).These input-output pairs form a training dataset for each user, enabling the LLM to be trained specifically for that user’s profile. By training the model on these personalized pairs, the LLM learns to generate tailored outputs that align with the user-specific information encoded in their profile. The LLM is trained on these pairs by minimizing the sequence-to-sequence cross-entropy loss \cite{NIPS2014_a14ac55a}. This loss function encourages the model to generate the target output \( y_i \) given the input \( x_i \).

There are various ways to implement the \( \textsc{convert} \) function depending on the structure of the user profile. If the user profile consists of input-output pairs, such as previous inputs from the user and the corresponding preferred outputs (e.g., rated by thumbs up or directly written by the user), these pairs can be directly used to train the model. In this case, the function \( \textsc{convert}(d_i) \) would simply map each document \( d_i \) to the corresponding input-output pair \( (x_i, y_i) \), which can then be used for training the LLM. Alternatively, when the user profile does not consist of explicit input-output pairs, such as when the profile contains previous comments or documents written by the user, these pairs can be automatically generated from each document in the profile. In such cases, text completion can be used as a task to teach the LLM the token distribution preferred by the user. The \( \textsc{convert} \) function can then generate input-output pairs by selecting parts of the user's previous text as the input and the next segment of text as the target output. This allows the model to learn patterns and preferences based on the user's writing style, tone, and content. Whenever the user profile contains explicit input-output pairs, we use those directly for training. If the profile does not contain such pairs, we define text completion as the training task. Specifically, we randomly select 10-20\% of the text as the input and use the remaining portion as the output. The conversion function for these tasks is summarized in Table \ref{tab:imp-convert}.

\begin{table*}
    \centering
    \caption{Statistics of the datasets within the LaMP benchmark \cite{lamp} with time-based data separation configuration used in our experiments in this paper. We use the time-based setting to study the effect of recency on the retrieval model's performance. Notably, this is the only setting where users are shared between the training and test sets, allowing us to evaluate PEFT methods.}
    \begin{adjustbox}{max width=\textwidth}    
        \begin{tabular}{l|ccc|cc|c|c}
            \toprule
            \textbf{Task}  & \textbf{\#train} & \textbf{\#dev} & \textbf{\#test} & \textbf{Input Length} & \textbf{Output Length} & \textbf{\#Profile Size} & \textbf{\#classes}\\
            \midrule
            {LaMP-1: Personalized Citation Identification}  & 6542 & 1500 & 1500 & 51.43 $\pm$ 5.70 & - & 84.15 $\pm$ 47.54 & 2 \\
            \midrule
            {LaMP-2: Personalized Movie Tagging}  & 5073 & 1410 & 1557 & 92.39 $\pm$ 21.95 & - & 86.76 $\pm$ 189.52 & 15 \\
            \midrule
            {LaMP-3: Personalized Product Rating}  & 20000 & 2500 & 2500 & 128.18 $\pm$ 146.25 & - & 185.40 $\pm$ 129.30 & 5 \\
            \midrule
            {LaMP-4: Personalized News Headline Generation} & 12500 & 1500 & 1800 & 29.97 $\pm$ 12.09 & 10.07 $\pm$ 3.10 & 204.59 $\pm$ 250.75 & - \\
            \midrule
            {LaMP-5: Personalized Scholarly Title Generation} & 14682 & 1500 & 1500 & 162.34 $\pm$ 65.63 & 9.71 $\pm$ 3.21 & 87.88 $\pm$ 53.63 & - \\
            \midrule
            {LaMP-6: Personalized Email Subject Generation}  & 4821 & 1250 & 1250 & 454.87 $\pm$ 889.41 & 7.37 $\pm$ 2.78 & 55.67 $\pm$ 36.32 & - \\
            \midrule
            {LaMP-7: Personalized Tweet Paraphrasing} & 13437 & 1498 & 1500 & 29.72 $\pm$ 7.01 & 16.96 $\pm$ 5.67 & 15.71 $\pm$ 14.86 & - \\ \bottomrule
        \end{tabular}
    \end{adjustbox}
    \label{tab:task-stats}
\end{table*}

\subsection{PEFT-RAG for Personalizing LLMs}

This approach integrates PEFT and RAG to enhance LLM personalization by leveraging their complementary strengths. PEFT facilitates efficient adaptation by fine-tuning a small number of additional parameters, reducing computational and memory overhead while allowing the model to capture user-specific preferences. However, PEFT alone has limitations, as it relies on static updates and may struggle to learn less common user preferences effectively \cite{mallen-etal-2023-trust}. RAG mitigates this issue by retrieving user-specific information at inference time, ensuring that personalization remains dynamic and contextually relevant without requiring extensive fine-tuning. By combining both approaches, the model benefits from long-term adaptation through fine-tuned parameters and real-time personalization through retrieval.

To do this, first, we train the LLM \(M\) on a user profile \(P_u\) using the method described in Section \ref{sec:peft}, which results in the personalized LLM $M_u$. Next, we apply the RAG personalization approach outlined in Section \ref{sec:rag}, denoted as:
\begin{equation}
    \label{eq:personal-rag}
    \hat{y} = M_{u}(\phi_{p}(x, R(\phi_{q}(x), k))
\end{equation}
where $M_u$ is a personalized model with PEFT as described in Section~\ref{sec:peft}, $R$ is a retriever, $\phi_q$ and $\phi_p$ are the query and prompt generation functions as shown in Table~\ref{tab:task-prompts}, and $k$ is the number of retrieved documents. The key difference from Equation \ref{eq:general-rag} is that here we first trains the LLM on the user-specific profile to get a personalized train LLM $M_u$ and learn the user preferences, the utilizing RAG as explained in Section~\ref{sec:rag} to further personalize the LLM.

\begin{table*}
    \centering
    \caption{Performance of different methods (RAG-based and PEFT-based) used for LLM personalization on the datasets in the LaMP benchmark. The results indicate that RAG methods are more effective than PEFT for personalizing LLMs. Moreover, combining RAG and PEFT achieves the best performance in LLM personalization.}
    \adjustbox{max width=\textwidth}{\begin{tabular}{l|l|c|cccc|cccc|cccc}
        \toprule
        \multirow{2}{*}{\textbf{Dataset}} & \multirow{2}{*}{\textbf{Metric}} & {\textbf{No}} & \multicolumn{4}{c|}{\textbf{PEFT Personalization}} & \multicolumn{4}{c|}{\textbf{RAG Personalization}} & \multicolumn{4}{c}{\textbf{PEFT-RAG Personalization}} \\
        & & \textbf{Personalization} & $r=8$ & $r=16$ & $r=32$ & $r=64$ & BM25 & Recency & Contriever & RSPG &  $r=8$ & $r=16$ & $r=32$ & $r=64$ \\
        \midrule

        \multirow{2}{*}{\shortstack[l]{{LaMP-1: Personalized}\\{Citation Identification}}} & \multirow{2}{*}{Accuracy $\uparrow$} & \multirow{2}{*}{0.502} & \multirow{2}{*}{0.502} & \multirow{2}{*}{0.502} & \multirow{2}{*}{0.504} & \multirow{2}{*}{0.506} & \multirow{2}{*}{0.626} & \multirow{2}{*}{0.622} & \multirow{2}{*}{0.636} & \multirow{2}{*}{\textbf{0.672}} & \multirow{2}{*}{0.670} & \multirow{2}{*}{0.668} & \multirow{2}{*}{0.671} & \multirow{2}{*}{0.671} \\
        & & & & & & & & & & & \\\midrule

        \multirow{2}{*}{\shortstack[l]{{LaMP-2: Personalized}\\{Movie Tagging}}} & Accuracy $\uparrow$ & 0.359 & 0.360 & 0.360 & 0.360 & 0.359 & 0.387 & 0.377 & 0.396 & 0.430 & 0.430 & \textbf{0.431} & 0.430 & 0.430 \\
        & F1 $\uparrow$ & 0.276 & 0.278 & 0.278 & 0.278 & 0.277 & 0.306 & 0.295 & 0.304 & 0.339 & 0.341 & \textbf{0.342} & 0.341 & 0.341 \\\midrule

        \multirow{2}{*}{\shortstack[l]{{LaMP-3: Personalized}\\{Product Rating}}} & MAE $\downarrow$ & 0.308 & 0.308 & 0.307 & 0.306 & 0.301 & 0.298 & 0.296 & 0.299 & 0.264 & 0.264 & 0.265 & 0.264 & \textbf{0.259} \\
        & RMSE $\downarrow$ & 0.611 & 0.607 & 0.607 & 0.602 & 0.600 & 0.611 & 0.605 & 0.616 & 0.568 & 0.568 & 0.570 & 0.564 & \textbf{0.562} \\\midrule

        \multirow{2}{*}{\shortstack[l]{{LaMP-4: Personalized}\\{News Headline Generation}}} & ROUGE-1 $\uparrow$ & 0.176 & 0.178 & 0.177 & 0.178 & 0.178 & 0.186 & 0.189 & 0.183 & 0.203 & 0.203 & \textbf{0.204} & \textbf{0.204} & 0.203 \\
        & ROUGE-L $\uparrow$ & 0.160 & 0.162 & 0.162 & 0.163 & 0.163 & 0.171 & 0.173 & 0.169 & 0.186 & 0.186 & 0.186 & \textbf{0.187} & 0.186 \\\midrule

        \multirow{2}{*}{\shortstack[l]{{LaMP-5: Personalized}\\{Scholarly Title Generation}}} & ROUGE-1 $\uparrow$ & 0.478 & 0.478 & 0.478 & 0.477 & 0.478 & 0.477 & 0.475 & \textbf{0.483} & 0.480 & 0.481 & 0.480 & 0.480 & 0.479 \\
        & ROUGE-L $\uparrow$ & 0.428 & 0.429 & 0.429 & 0.428 & 0.428 & 0.427 & 0.426 & \textbf{0.433} & 0.429 & 0.431 & 0.431 & 0.431 & 0.431 \\\midrule

        \multirow{2}{*}{\shortstack[l]{{LaMP-6: Personalized}\\{Email Subject Generation}}} & ROUGE-1 $\uparrow$ & 0.335 & 0.342 & 0.342 & 0.341 & 0.343 & 0.412 & 0.403 & 0.401 & 0.433 & 0.436 & 0.436 & 0.436 & \textbf{0.437} \\
        & ROUGE-L $\uparrow$ & 0.319 & 0.325 & 0.326 & 0.325 & 0.326 & 0.398 & 0.389 & 0.386 & 0.418 & \textbf{0.422} & \textbf{0.422} & \textbf{0.422} & 0.421 \\\midrule

        \multirow{2}{*}{\shortstack[l]{{LaMP-7: Personalized}\\{Tweet Paraphrasing}}} & ROUGE-1 $\uparrow$ & 0.449 & 0.449 & 0.449 & 0.449 & 0.449 & 0.446 & 0.444 & 0.440 & \textbf{0.461} & 0.460 & 0.460 & 0.460 & 0.460 \\
        & ROUGE-L $\uparrow$ & 0.396 & 0.397 & 0.397 & 0.396 & 0.396 & 0.394 & 0.393 & 0.390 & \textbf{0.409} & \textbf{ 0.409} & \textbf{0.409} & 0.408 & \textbf{0.409} \\ \bottomrule
        
    \end{tabular}}
    
    \label{tab:main-results}
\end{table*}

\section{Experiments}

This section details our findings about comparing RAG and PEFT for privacy-preserving personalization of LLMs.

\subsection{Experimental Setup}

\subsubsection*{\textbf{Datasets \& Tasks}}

Our experiments utilize the LaMP benchmark \cite{lamp}, which is specifically designed to evaluate LLM personalization. Each instance in this benchmark represents a user and includes an input prompt, an expected output, and a user profile comprising structured or unstructured documents. LaMP covers seven personalized tasks, including three text classification tasks and four text generation tasks. Table \ref{tab:task-stats} provides detailed dataset statistics. We adopt the time-based configuration of LaMP, as it ensures that the same users appear in both the training and test sets, enabling effective model training on user profiles for the PEFT approach. For evaluation, following previous work \cite{lamp, rspg}, we use accuracy for binary classification (LaMP-1), accuracy and F1 score for categorical classification (LaMP-2), and MAE and RMSE for ordinal classification (LaMP-3). For text generation tasks (LaMP-4 to LaMP-7), we assess performance using ROUGE-1 and ROUGE-L \cite{rouge}.

\subsubsection*{\textbf{RAG Pipeline Configuration.}}

For personalizing LLMs using retrieval-augmented generation, we adopt the experimental setup from \citet{lamp} and \citet{rspg}. Specifically, we utilize the BM25 \cite{bm25} retrieval model, implemented in the \textit{rank\_bm25} library\footnote{Available at \url{https://github.com/dorianbrown/rank\_bm25}}, along with Contriever\footnote{Available at \url{https://huggingface.co/facebook/contriever}} \cite{contriever}, Recency-based retrieval, and RSPG \cite{rspg}. In all experiments, we retrieve \( k = 4 \) documents from the user profile to personalize the LLM. Following \citet{lamp}, we use FlanT5-XXL\footnote{Available at \url{https://huggingface.co/google/flan-t5-xxl}} \cite{flant5}, a model with 11 billion parameters, as our base LLM. The model is configured with an input length of 512 tokens and an output length of 128 tokens. For text generation, we employ beam search \cite{beam-search} with a beam size of 4. All experiments are conducted using the Hugging Face library\footnote{Available at \url{https://huggingface.co/}} \cite{wolf-etal-2020-transformers}.

\subsubsection*{\textbf{PEFT Pipeline Configuration.}}

To train the LLMs for each user, we utilize the PEFT library\footnote{Available at \url{https://huggingface.co/docs/peft/en/index}}. Each model is trained for 50 epochs on the corresponding user profile with a learning rate of $5 \times 10^{-4}$, applying a linear scheduler with a warm-up phase covering 5\% of the total training steps. We use the Adam optimizer \cite{adam} with a weight decay of $10^{-4}$, and a batch size of 16 is achieved through gradient accumulation. LoRA is employed with a dropout rate of 0.1 and a scaling factor $\alpha = 32$, applied to all key, query, and value projections in the transformer \cite{transformer}. Following \citet{lamp}, we use FlanT5-XXL\footnote{Available at \url{https://huggingface.co/google/flan-t5-xxl}} \cite{flant5}, a model with 11 billion parameters, as our base LLM. The model is configured with an input length of 512 tokens and an output length of 128 tokens. For text generation, we employ beam search \cite{beam-search} with a beam size of 4. The experiments are conducted on up to 32 Nvidia A100 GPUs (80GB VRAM) with 128GB RAM over a period of up to 7 days. In total, our experiments consumed over 10,000 GPU hours. To mitigate computational costs, we train an LLM only for users present in the test set rather than for all users in the benchmark. As a result, 37,560 LoRA adapters were trained, occupying approximately 18 TB of disk space.

\subsection{Main Findings}

\subsubsection*{\textbf{RQ1: How do PEFT- and RAG-based approaches perform for LLM personalization?}}

The results of both PEFT-based and RAG-based personalization methods, along with the non-personalized baseline, are reported in Table~\ref{tab:main-results}. Our findings about this question in Table~\ref{tab:main-results} indicate that incorporating PEFT improves performance over non-personalized LLMs in 5 out of the 7 evaluated datasets, highlighting its effectiveness in leveraging parameter-efficient tuning for personalization. Similarly, the RAG approach demonstrates consistent performance improvements across all datasets, suggesting that retrieval-based personalization is a robust method for enhancing LLM outputs. When comparing PEFT with RAG, our analysis in Table~\ref{tab:main-results} reveals that RAG is a more effective personalization strategy. As shown in Table~\ref{tab:main-results}, PEFT leads to a modest average performance improvement of 1.07\% over non-personalized LLMs, whereas the RAG-based approach yields a significantly higher average improvement of 14.92\%. This substantial performance gap underscores the advantage of retrieval-augmented generation in adapting LLMs to personalized contexts.

It is worth noting that different retrieval models in Table~\ref{tab:main-results} exhibit varying levels of effectiveness in improving performance. Among them, RSPG \cite{rspg}, which dynamically selects the most suitable retrieval model for each instance, consistently outperforms all other retrieval-based methods, achieving the highest overall performance. This highlights the importance of adaptive retrieval mechanisms in optimizing personalization. Additionally, we observe that the rank parameter $r$ in PEFT also influences model performance, though its impact varies across datasets. Specifically, on LaMP-1, LaMP-3, and LaMP-6, increasing $r$ leads to noticeable performance improvements, suggesting that a higher rank enables better adaptation to personalized contexts in these tasks. However, for other datasets, the effect of adjusting $r$ is less pronounced, indicating that the benefits of increasing parameter efficiency may be task-dependent.


\subsubsection*{\textbf{RQ2: How does the combination of PEFT and RAG impact the personalization performance?}}

To answer this research question, we integrate the best-performing retrieval model from Table~\ref{tab:main-results} with each user's personalized LLM, trained using PEFT, to perform RAG personalization with PEFT. This approach aims to leverage both retrieval-augmented generation and parameter-efficient fine-tuning to maximize personalization effectiveness. The results of this experiment are presented in Table~\ref{tab:main-results}. Our findings indicate that combining RAG with PEFT yields performance improvements over the standard RAG approach in 4 out of the 7 evaluated tasks. This suggests that fine-tuning a personalized LLM alongside retrieval-based augmentation allows the model to better adapt to user-specific preferences and contextual nuances. Moreover, this combined approach results in an overall 15.98\% improvement over the non-personalized LLM, surpassing standard RAG personalization by an additional 0.44\% in relative performance gain. Although the improvement over RAG alone is modest, it demonstrates that incorporating PEFT can provide additional benefits, particularly in scenarios where retrieval alone may not fully capture personalization needs. These results highlight the potential of combining PEFT with RAG as a viable strategy for further enhancing LLM personalization, offering a balanced approach that benefits from both retrieval-augmented generation and parameter-efficient adaptation.


\begin{figure*}
\centering
\includegraphics[width=\linewidth]{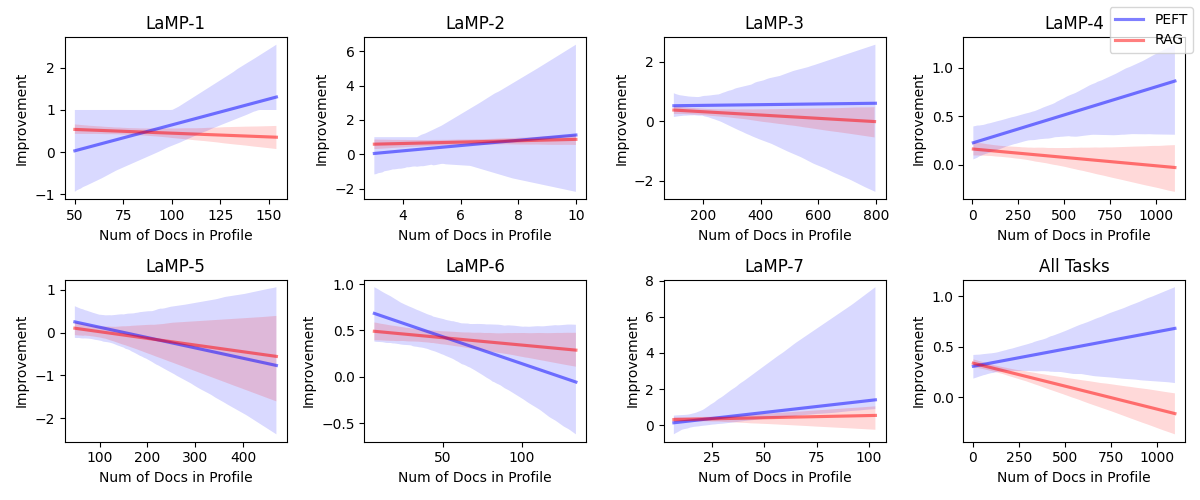}
        \caption{Correlation (with 95\% confidence interval) between profile item count and performance improvement of \textcolor{blue}{PEFT}- and \textcolor{red}{RAG}-based personalization in comparison with no personalization on the tasks in the LaMP benchmark. The results indicate a positive correlation between the improvement of PEFT over non-personalized LLMs and the profile size, suggesting that the limited amount of personalized data per user contributes to PEFT's underperformance. Conversely, RAG exhibits a negative correlation with profile size, indicating that as the profile grows, the retrieval model faces challenges in retrieving relevant information for effective LLM personalization.}
    \label{fig:doc-count-performance}
\end{figure*}

\subsubsection*{\textbf{RQ3: How does profile size affect performance?}}

We create a regression plot\footnote{\url{https://seaborn.pydata.org/generated/seaborn.regplot.html}} that visualizes the relationship between the number of documents in a user's profile and the relative improvement achieved by the best-performing personalized LLM using PEFT- and RAG-based personalization compared to the non-personalized LLM. Here, we define improvement as 1 when there is a performance gain over the non-personalized baseline and -1 when there is no gain. To ensure that our analysis focuses on meaningful variations, we exclude users who experience no change in performance. The resulting plot is shown in Figure~\ref{fig:doc-count-performance}. This figure reveals that in 5 out of the 7 datasets, there is a positive correlation between the number of items in a user’s profile and the performance improvement achieved through PEFT-based personalization. This suggests that as more user-specific data becomes available for fine-tuning, the model is better able to adapt to individual preferences and generate more personalized responses. However, in the case of the LaMP-5 task, we observe either a negative or zero correlation. One possible explanation for this anomaly is the nature of the user profiles, which consist of abstracts from papers authored by the user. Since many of these papers are collaborative efforts, users with larger profiles tend to be senior researchers who may not have been directly involved in the writing process. Our analysis further supports this hypothesis, as we found that in 94\% (17 out of 18) of the cases where performance dropped, the user was not the primary author on most papers in their profile. This suggests that training on such collaborative data is less effective for personalizing the LLM to the user's actual writing style and preferences.

When aggregating results across all tasks, we observe an overall positive correlation between PEFT-based performance improvement and the number of items in a user’s profile. In contrast, we find a negative correlation between performance improvement using RAG-based personalization and the non-personalized baseline. This indicates that as the size of a user’s profile increases, retrieval models struggle to accurately identify and retrieve the most relevant documents for personalization, potentially leading to diminished gains. These findings provide insight into the strengths and limitations of PEFT- and RAG-based personalization. While PEFT benefits from having more user-specific training data, its effectiveness is constrained when the available data is insufficient for robust adaptation. On the other hand, RAG’s performance appears to degrade as the profile size grows, likely due to challenges in retrieval quality. This suggests that one of the primary reasons PEFT does not consistently outperform RAG for personalizing LLMs is the limited amount of training data per user, which restricts the model’s ability to effectively learn individual preferences.

\subsubsection*{\textbf{RQ4: How does data presence in training corpus affect performance?}}

Another notable observation from Table~\ref{tab:main-results} is that PEFT achieves the highest performance gains on the LaMP-6 task compared to other evaluated tasks. We hypothesize that this is due to the nature of the dataset and the pretraining corpus of FlanT5 \cite{flant5}. Since FlanT5 is trained on publicly available datasets, it has not been exposed to private datasets such as LaMP-6, which contains personally identifiable information (PII) from the Avocado \cite{avocado} corpus. As a result, the model encounters this corpus for the first time during fine-tuning, allowing it to learn new, task-specific patterns more effectively. In contrast, other datasets used in our experiments are primarily derived from public sources that were likely included in FlanT5’s pretraining corpus. This means that the model has already seen similar data during pretraining, leading to relatively smaller performance improvements when fine-tuned using PEFT. These findings suggest that applying PEFT to private user data can lead to considerable performance gains, particularly in scenarios where the LLM has not previously encountered the data. This underscores the potential of PEFT for enhancing personalization in cases where user-specific data is distinct from publicly available corpora.







\section{Limitations of RAG and PEFT for LLM Personalization}

This section discusses the limitations of personalizing LLMs using RAG and PEFT. In addition, it outlines future research directions that could help mitigate these challenges, paving the way for more effective personalization techniques that can be deployed in real-world systems that serve millions of users.

\subsection{Resource Intensivity}

Personalizing large language models, particularly using PEFT with LoRA, can be computationally demanding. Training models with LoRA requires substantial resources, leading to increased costs and extended training times, which can hinder scalability in resource-constrained environments. Due to these limitations, our experiments in this paper were conducted on just one single large language model, FlanT5-XXL \cite{flant5}, which has 11 billion parameters, following \citet{lamp}. While investigating personalization across multiple LLM architectures would provide valuable insights, the computational costs are prohibitively high.  

In this work, we utilized over 10,000 hours of A100 GPU computation for training and experimentation. Based on the average figures reported by \citet{10.1145/3531146.3533234}, running these experiments on cloud-based GPU providers would result in the generation of at least 400 kilograms of CO2 emissions. Since our experiments were conducted locally, where energy efficiency may be lower, the actual carbon footprint could be even higher. This underscores the environmental impact of large-scale LLM experiments and the challenges of scaling personalized LLMs to millions of users. Expanding this study to include multiple LLMs would significantly increase the computational burden, making it infeasible at this time.  

Beyond the costs associated with training, storing personalized large language models presents another challenge. If we assume on average each adapter requires 200 MB of disk space, a platform with 100 million users would require approximately 20 PB of storage just for storing adapters. Such large storage requirements pose a significant challenge for deploying personalized LLMs at scale. Addressing these limitations—both computational and storage-related—is essential for the practical deployment of personalized generative AI systems. Future research should explore solutions such as parameter-efficient methods, adaptive storage mechanisms, or dynamic retrieval approaches to mitigate these constraints and facilitate real-world applications of personalized large language models.

\subsection{Adapter Loading and Retrieval Latency}

In our comparison between RAG and PEFT methods, adapter loading and retrieval latency emerge as critical factors influencing overall system performance. Each approach presents unique challenges that impact deployment efficiency and real-time usability. For RAG-based models, retrieval latency is a significant concern. Querying external databases and fetching relevant information introduces time costs that can affect system responsiveness. High retrieval latency can be particularly problematic for real-time applications, where rapid responses are essential. Additionally, managing retrieval efficiency involves challenges such as effective indexing strategies, optimizing search algorithms, and handling large-scale corpora, all of which can exacerbate latency issues. If retrieval is slow or inaccurate, it can degrade the model’s effectiveness, as the retrieved information may not be timely or contextually relevant.

Conversely, PEFT-based approaches rely on adapting pre-trained models by loading user-specific adapters. While PEFT is more resource-efficient than full fine-tuning, adapter loading introduces its own overhead, especially in large-scale deployments. Initializing multiple adapters or switching between them dynamically can impose computational burdens that affect system throughput. This challenge is particularly pronounced in scenarios requiring frequent updates or real-time interactions, where loading delays may introduce unwanted lag. Additionally, when working with millions of personalized adapters, memory management becomes critical factors for maintaining scalability. Thus, while RAG-based personalization suffers from retrieval latency, PEFT-based personalization is constrained by adapter loading overhead. Addressing these bottlenecks—whether through optimized retrieval pipelines for RAG or efficient adapter---loading mechanisms for PEFT---is essential for enabling seamless personalization in LLM-powered applications. Future work should explore solutions such as cached retrieval strategies, lightweight indexing, model compression techniques, or dynamic adapter management to improve both efficiency and scalability in personalized LLM systems.


\section{Conclusion}

This paper presents the first systematic comparison of retrieval-augmentation and parameter-efficient fine-tuning for personalizing LLMs in a privacy-preserving setting. Additionally, we propose a hybrid approach that combines both methods to enhance LLMs' ability in personalized text generation. Our experiments on the LaMP benchmark demonstrate that both RAG and PEFT improve LLM performance on personalized tasks. However, RAG significantly outperforms PEFT in this setting. Moreover, we find that the best results are achieved when RAG and PEFT are combined, leading to a 15.98\% improvement over the performance of a non-personalized LLM on the LaMP benchmark tasks. Further analysis reveals a positive correlation between the number of documents available for each user and the performance gain from PEFT-based personalization. This finding suggests that the primary limitation of PEFT is the lack of sufficient per-user training data, which restricts its effectiveness when used alone. Finally, we show that PEFT is particularly effective when trained on private user data, as the LLM has not encountered this data during pretraining. This highlights PEFT’s potential for leveraging private information in a way that standard pretraining and retrieval-augmented approaches cannot.


\section*{Acknowledgment}
The authors would like to thank Mohammad Aliannejadi for his feedback on a draft of this paper. This work was supported in part by the Center for Intelligent Information Retrieval, in part by Google, in part by Lowe’s, and in part by Microsoft. Any opinions, findings and conclusions or recommendations expressed in this material are those of the authors and do not necessarily reflect those of the sponsor.

\balance
\bibliographystyle{ACM-Reference-Format}
\bibliography{XX-references}

\appendix

\end{document}